\documentclass[preprint,12pt]{elsarticle}

\usepackage{amssymb}
\usepackage{amsmath}
\usepackage{times}
\usepackage{soul}
\usepackage[table]{xcolor}
\usepackage{url}
\usepackage[utf8]{inputenc}
\usepackage[small]{caption}
\usepackage{graphicx}
\usepackage{diagbox} 
\usepackage{amsthm}
\usepackage{booktabs}
\usepackage{algorithm}
\usepackage{algpseudocode}
\usepackage[switch]{lineno}
\usepackage{amssymb}
\usepackage{subcaption}
\usepackage{enumitem}
\usepackage{multirow}
\usepackage{subfiles} 
\usepackage{pdfpages}
\usepackage{booktabs}    % 提供toprule/midrule/bottomrule等专业表格线
\usepackage{booktabs}   % 支撑\toprule/\midrule/\bottomrule/\cmidrule
\usepackage{multirow}   % 支撑\multirow命令
\usepackage{makecell}   % 支撑\makecell命令（单元格内换行）
\usepackage{siunitx}     % 用于数字列的对齐（S类型列）
\usepackage{xcolor}      % 支持单元格背景色（gray!10）
\usepackage{caption}     % 优化表格标题格式

\usepackage[margin=1in]{geometry}
\usepackage{amsmath,amssymb,amsfonts}
\usepackage{mathtools}
\usepackage{bm}

\journal{Knowledge-Based Systems}

\begin{document}

\begin{frontmatter}



\title{FedPSA: Modeling Behavioral Staleness in
Asynchronous Federated Learning} %% Article title

\author[1,2,3]{Chaoyi Lu\fnmark[label2]}\ead{chaoyi@stu.xjtu.edu.cn} %% Author name
\author[1]{Yiding Sun\fnmark[label2]}\ead{sunyiding@stu.xjtu.edu.cn}
\author[1]{Zhichuan Yang}\ead{zhichuan@stu.xjtu.edu.cn}
\author[1]{Jinqian Chen}\ead{chenjinqian@stu.xjtu.edu.cn}
\author[3]{Dongfu Yin\cormark[cor1]}\ead{yindongfu@gml.ac.cn}
\author[1,2]{Jihua Zhu\cormark[cor1]}\ead{zhujh@xjtu.edu.cn}
\fntext[label2]{Equal Contribution}
\cortext[cor1]{Corresponding Author}
\affiliation[1]{organization={School of Software Engineering},
                addressline={Xi'an Jiaotong University}, 
                city={Xi'an},
%               citysep={}, % Uncomment if no comma needed between city and postcode
                postcode={710049}, 
                country={China}}
\affiliation[2]{organization={State Key Laboratory of Integrated Services Networks},
                addressline={Xidian University}, 
                city={Xi'an},
%               citysep={}, % Uncomment if no comma needed between city and postcode
                postcode={710071}, 
                country={China}}
\affiliation[3]{
                addressline={Guangdong Laboratory of Artificial Intelligence and Digital Economy(SZ)}, 
                city={Shenzhen},
%               citysep={}, % Uncomment if no comma needed between city and postcode
                postcode={ 518107}, 
                country={China}}

%% Abstract
\begin{abstract}
Asynchronous Federated Learning (AFL) has emerged as a significant research area in recent years. By not waiting for slower clients and executing the training process concurrently, it achieves faster training speed compared to traditional federated learning. However, due to the staleness introduced by the asynchronous process, its performance may degrade in some scenarios. Existing methods often use the round difference between the current model and the global model as the sole measure of staleness, which is coarse-grained and lacks observation of the model itself, thereby limiting the performance ceiling of asynchronous methods. In this paper, we propose FedPSA (Parameter Sensitivity-based Asynchronous Federated Learning), a more fine-grained AFL framework that leverages parameter sensitivity to measure model obsolescence and establishes a dynamic momentum queue to assess the current training phase in real time, thereby adjusting the tolerance for outdated information dynamically. Extensive experiments on multiple datasets and comparisons with various methods demonstrate the superior performance of FedPSA, achieving up to 6.37\% improvement over baseline methods and 1.93\% over the current state-of-the-art method.
\end{abstract}

\begin{graphicalabstract}
\begin{figure*}[h]
    \centering
    \begin{minipage}{1.0\textwidth}
        \includegraphics[width=\linewidth]{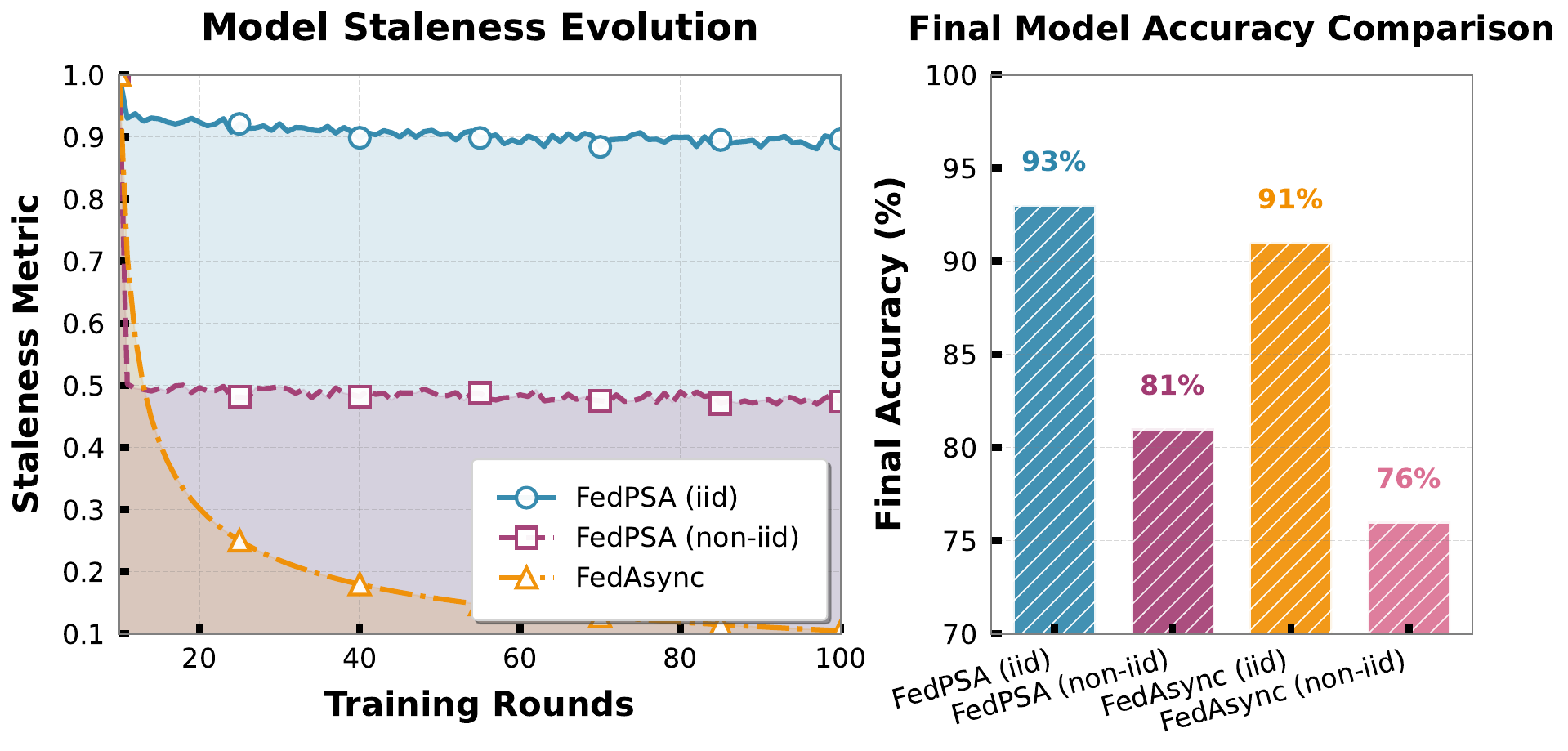}
    \end{minipage}%
\end{figure*}
    
\textbf{Comparison of weighting coefficients and final accuracy between FedPSA and FedAsync.} Traditional methods overlook the details during aggregation, leading to poor final performance.

\end{graphicalabstract}

\begin{highlights}
\item We propose FedPSA, a novel asynchronous federated learning framework that moves beyond round-gap-based staleness by introducing a behavioral staleness metric derived from parameter-sensitivity patterns, enabling the server to directly assess the compatibility between client updates and the current global training dynamics.
\item We design an efficient and stage-adaptive aggregation scheme: clients compute sensitivity on a shared calibration batch, compress it via random projection sketching, and the server combines the resulting cosine-similarity signal with a momentum-queue-based training thermometer to dynamically adjust the sharpness of softmax weighting across training stages.
\item We validate FedPSA via extensive experiments and ablation studies on multiple benchmarks. The results show that FedPSA achieves strong performance under both IID and non-IID data distributions, and is resilient to heterogeneous system latency.
\end{highlights}

%% Keywords

\end{frontmatter}

\section{Introduction}
\label{sec:intro}
Over the course of recent years, the increasing emphasis on information security has drawn researchers' attention to a novel machine learning paradigm known as Federated Learning (FL) \cite{9084352,10.1561/2200000083}. As a privacy-preserving machine learning method, it replaces data sharing with model parameter sharing, thereby avoiding data leakage while maintaining training effectiveness \cite{Zhang2025,Wang2025}. To coordinate multiple clients for collaborative training, the traditional FL process operates as follows \cite{pmlr-v54-mcmahan17a}: First, a server is established to collect and broadcast various types of information during the training process. In each round, the server selects a subset of clients and broadcasts the current global model to them, instructing them to train the model using their local data~\cite{LU2025114632,USMAN2025114019,LI2026114972,ZHANG2026115335}. The server then waits for all selected clients to complete their training and upload their updated models. Afterward, it aggregates all the received client models to form a new global model for the next round. This process repeats until a predefined number of rounds is reached or other exit conditions are met. By transmitting model parameters instead of the training data itself, FL effectively ensures data security.

The drawbacks of traditional FL are also evident: since each round of training requires waiting for all selected clients to complete their training, even if just one client responds slowly in a given round, it can significantly slow down the entire training progress. This issue is referred to as the ``straggler problem" \cite{10.5555/3666122.3669324,10.1145/3637528.3671979,NEURIPS2021_076a8133}, which severely limits the broader applicability of traditional FL, making it difficult to deploy in scenarios with significant disparities in device performance \cite{Singh2025}. To address this problem, researchers have recently proposed an Asynchronous Federated Learning (AFL) framework \cite{xie2020asynchronousfederatedoptimization,wang2024tackling}. By maintaining a pool of participating clients above a specified concurrency threshold, rather than designating and waiting for a fixed set of clients to complete training in each round, this framework effectively prevents the emergence of stragglers.

However, the asynchronous nature of AFL also raises new concerns. In this framework, the server does not wait for all clients to finish training before performing aggregation. By the time a client completes its training and uploads its model, the global model on the server may have been updated many times. This creates a version difference between the client's model and the server's model. This version gap is referred to as staleness \cite{10.1145/3527621,10561520,10208323}. Since the uploaded models from clients can be outdated, the information they contribute is less valuable. Although AFL performs updates much more frequently, the lower quality of each update often results in poorer overall training performance.

To mitigate the impact of staleness in AFL, existing methods can be broadly categorized into two directions: 1) Establishing a buffer on the server to receive gradients and performing aggregation only when the buffer is full \cite{Niu2026,9093123}. This method reduces the frequency of global model updates from aggregating every time gradients are uploaded to aggregating only when the buffer is full, effectively decreasing the negative impact of staleness. 2) Reducing the weight of outdated information \cite{xie2020asynchronousfederatedoptimization,9484767}. Most methods in this category still rely solely on version differences to measure staleness, lacking attention to the detailed information of the models themselves, which in fact limits the performance of AFL. Therefore, existing methods still struggle to fully leverage the advantages of AFL.

In this paper, we propose a novel AFL framework, FedPSA, which further unleashes the potential of AFL by observing and leveraging the behavioral information of the models themselves. We define behavioral staleness as a measure of discrepancy between client and global model behaviors, quantified via parameter sensitivity patterns. Parameter sensitivity is derived from each model itself and can effectively quantify the conflict level between models, helping to determine whether they can collaborate smoothly. To enhance the exploratory nature of the algorithm during the early training stages, FedPSA also incorporates a dynamic queue to store the recent training momentum uploaded by clients. By averaging the momentum stored in the queue, FedPSA estimates the current training phase. Consequently, it can accommodate larger variations in parameter sensitivity during the exploratory early stages, while imposing stricter constraints on sensitivity differences as the model approaches convergence. Our contributions are as follows:

\begin{itemize}

\item We propose FedPSA, a novel asynchronous federated learning framework that moves beyond round-gap-based staleness by introducing a behavioral staleness metric derived from parameter-sensitivity patterns, enabling the server to directly assess the compatibility between client updates and the current global training dynamics.

\item We design an efficient and stage-adaptive aggregation scheme: clients compute sensitivity on a shared calibration batch, compress it via random projection sketching, and the server combines the resulting cosine-similarity signal with a momentum-queue-based training thermometer to dynamically adjust the sharpness of softmax weighting across training stages.

\item We validate FedPSA via extensive experiments and ablation studies on multiple benchmarks. The results show that FedPSA achieves strong performance under both IID and non-IID data distributions, and is resilient to heterogeneous system latency.
\end{itemize}

\section{Related Work}

\subsection{Traditional Federated Learning}
Federated learning was originally formulated as a synchronous distributed optimization paradigm, where a central server coordinates a subset of clients to perform local updates and aggregates them only after all selected clients have finished their computations. The seminal FedAvg algorithm~\cite{pmlr-v54-mcmahan17a} averages local model parameters from participating clients to update the global model, and has become the \emph{de facto} baseline for synchronous FL. To better handle data heterogeneity across clients, FedProx~\cite{li2020federatedoptimizationheterogeneousnetworks} introduces a proximal term that penalizes deviations from the global model during local training, while SCAFFOLD~\cite{10.5555/3524938.3525414} leverages control variates to correct client drift in non-IID scenarios. 

Subsequent work further improves robustness and accuracy under heterogeneous data. For example, FedCM~\cite{xu2021fedcmfederatedlearningclientlevel} exploits historical gradient information to reduce aggregation bias, and bMOM~\cite{10.1007/978-3-031-25158-0_19} adopts a bootstrap median-of-means strategy for robust aggregation against outliers. Despite their success, these methods all follow the synchronous “round-based” protocol: in each communication round, the server must wait for all selected clients to return their updates before proceeding. As a consequence, the overall training speed is dominated by the slowest devices, leading to the well-known ``straggler problem" and motivating the development of AFL.

\subsection{Asynchronous Federated Learning and Staleness Modeling}
AFL alleviates the ``straggler problem" by allowing the server to update the global model whenever a client upload arrives, without waiting for other clients. In one of the earliest AFL frameworks, FedAsync~\cite{xie2020asynchronousfederatedoptimization}, the server assigns aggregation weights according to the staleness of each update, which is defined as the version (or round) gap between the local model and the current global model. More recent algorithms refine this idea. FedASMU~\cite{10.1609/aaai.v38i12.29297} requires clients to pull the latest global model before local training to reduce bias, while FedFa~\cite{10.24963/ijcai.2024/584} maintains a queue on the server and discards outdated updates when the queue overflows. 

Another influential line of work introduces buffering mechanisms to mitigate the negative effect of stale updates. Buffer-based AFL methods~\cite{Niu2026,9093123} maintain a server-side buffer that temporarily accumulates multiple client updates and triggers aggregation only when the buffer is full. This design reduces the frequency of global model updates compared with aggregating every single incoming update, and has been shown to improve convergence stability in highly asynchronous environments. Beyond supervised learning, AFL has also been extended to more complex settings. Asyn2F~\cite{Cao2024Asyn2FAA} proposes bidirectional model aggregation between clients and the server, momentum-approximation techniques~\cite{Yu2024MomentumAI} enable scalable private AFL over large client pools, and AFedPG~\cite{DBLP:journals/corr/abs-2404-08003} adapts asynchronous ideas to policy-gradient reinforcement learning.

Despite these advances, the way staleness is quantified in AFL remains largely time-based. In most existing methods, the aggregation weight of a client update is modeled as a function of the iteration or version gap between the client and the server, \emph{e.g.}, through monotonically decreasing functions of the round difference. This treatment implicitly assumes that all updates with the same time gap should have the same importance, and ignores the behavioral information of the models themselves. As a result, these methods cannot distinguish, for example, between two equally delayed updates generated at different training stages or under different local data distributions. Our work moves beyond this purely temporal view and proposes a behavioral-aware staleness metric based on parameter sensitivity.

\subsection{Model Parameter Sensitivity and Behavior-Aware Aggregation}
Parameter sensitivity analysis has a long history in machine learning and neural networks. In general, it studies how perturbations to individual parameters or input features affect the model loss, and is widely used to improve interpretability, robustness, and pruning efficiency~\cite{scholbeck2024positionpaperbridginggap}. Classical methods such as the Lek-profile~\cite{LEK199639} systematically vary inputs or parameters to reveal nonlinear dependencies between explanatory and response variables. More recent work leverages data-driven perturbation strategies to explore high-dimensional input spaces and quantify the importance of individual or grouped features in complex engineering applications~\cite{TUNKIEL2020107630}.

In the context of FL, parameter sensitivity has been introduced mainly for model compression and personalization. For example, some methods adaptively interpolate client models based on parameter sensitivity and reduce the update frequency of highly sensitive parameters, thereby mitigating client drift in traditional FL~\cite{10204999}. FedCAC~\cite{Wu2023FedCAC} uses parameter sensitivity to determine collaboration relationships among clients, improving personalized FL by sharing only those parameters that are less sensitive to local objectives. These methods demonstrate that sensitivity is a powerful tool for capturing parameter importance and client heterogeneity.

However, existing sensitivity-based methods are still designed for synchronous or personalized FL and do not address the staleness issue in asynchronous settings. To the best of our knowledge, no prior work uses parameter sensitivity to quantify staleness and to directly control the aggregation weights in AFL. In contrast, FedPSA introduces a sensitivity-driven, behavior-aware staleness metric: it compares the sensitivity patterns of client and server models (on a shared calibration batch), compresses them via random projections into low-dimensional sketches, and uses their cosine similarity as a proxy for behavioral freshness. Combined with a training thermometer that adapts the tolerance to behavioral discrepancies across different training stages, this design enables a finer-grained and more informative staleness modeling than traditional schemes based solely on time gaps or version gaps.

\section{Preliminaries}
\label{sec:prelim}

We consider a standard FL setup with $n$ clients, indexed by $i \in \{1,\dots,n\}$. Client $i$ holds a private dataset $\mathcal{D}_i = \{(x_{i,j}, y_{i,j})\}_{j=1}^{|\mathcal{D}_i|}$, where $x_{i,j}$ denotes the input features and $y_{i,j}$ the corresponding label. Due to privacy or regulatory constraints, clients cannot share their raw data with a central entity. Instead, a trusted server coordinates model training by exchanging model parameters or updates with the clients.

The goal is to collaboratively learn a global model parameter $w \in \mathbb{R}^d$ that minimizes the following weighted empirical risk:
\begin{equation}
    F(w)
    = \sum_{i=1}^n p_i f_i(w),
\end{equation}
where
\(
    f_i(w) = \frac{1}{|\mathcal{D}_i|} \sum_{(x,y)\in\mathcal{D}_i} \ell(w; x,y)
\)
is the local loss on client $i$, $\ell(\cdot)$ is the task-specific loss function, and
\(
    p_i = \frac{|\mathcal{D}_i|}{\sum_{j=1}^n |\mathcal{D}_j|}
\)
reflects the relative data size of client $i$.

In synchronous FL, the server proceeds in communication rounds: it broadcasts the current global model to a subset of clients, waits until all selected clients finish local training, and then aggregates their updates to obtain the next global model~\cite{pmlr-v54-mcmahan17a}. This round-based protocol ensures that all aggregated updates are computed from the same global model, but also makes the overall training speed sensitive to the slowest clients.

AFL removes this strict synchronization requirement. Let $w^t$ denote the global model at (logical) iteration $t$, and suppose client $i$ performs local training starting from an older global model $w^{t-\tau_i}$, where $\tau_i \geq 0$ is the version gap between the local and current global models. After local training, client $i$ uploads an update $\Delta w_i^{t-\tau_i}$ to the server, and a generic asynchronous update can be written as
\begin{equation}
    w^{t+1}
    = w^t + \alpha_i \,\Delta w_i^{t-\tau_i},
    \label{eq:generic_async_update}
\end{equation}
where $\alpha_i$ is an aggregation weight that typically depends on the staleness $\tau_i$. Most existing AFL methods model staleness purely through this time or version-gap and choose $\alpha_i$ as a decreasing function of $\tau_i$, without inspecting the behavioral information of the models themselves. As discussed in Section~\ref{sec:motivation}, such a time-based metric is often too coarse to capture the true quality of client updates at different training stages and under heterogeneous data distributions.

\section{Motivation}
\label{sec:motivation}
Most existing asynchronous FL methods quantify model staleness solely by the round gap between a client model and the latest global model, \emph{i.e.}, using a scalar function $s(\tau)$ of the iteration difference $\tau$. The resulting aggregation weights depend only on $\tau$, implicitly assuming that updates with the same round gap should have the same importance. This simplification ignores several key factors.

First, the same round gap can correspond to very different degrees of model drift at different training stages. In the early phase, when the global model is still far from convergence, a stale update may deviate significantly from the current global model. In the late phase, when the model is close to convergence, an update with the same round gap may be almost indistinguishable from the latest global model. However, a purely round-gap-based staleness metric assigns them identical weights, regardless of their actual behavioral discrepancy.

Second, for two client models with the same round gap, their impact on the global model can still be very different under heterogeneous data. Intuitively, given the same $\tau$, the update from a client with a larger distribution shift should receive a smaller aggregation weight. If both updates are aggregated with equal weights, the global model will be biased towards clients whose local data diverge more from the global data distribution.

These examples illustrate a broader problem: when training stages differ and data distributions are heterogeneous, using a single scalar function of the iteration difference as the staleness metric inevitably limits the effectiveness of training. This motivates us to move beyond coarse, time-only notions of staleness and seek a metric that captures the behavioral discrepancy between models.

In this work, we propose to measure staleness from the perspective of parameter sensitivity. Parameter sensitivity reflects how critical each parameter is to the model's loss, thus providing fine-grained, parameter-level information about the model state. By comparing sensitivity patterns between client and server models, we can obtain a behavioral-aware notion of staleness that goes beyond mere iteration differences.

\begin{figure}[t]
    \centering
    \includegraphics[width=0.5\textwidth]{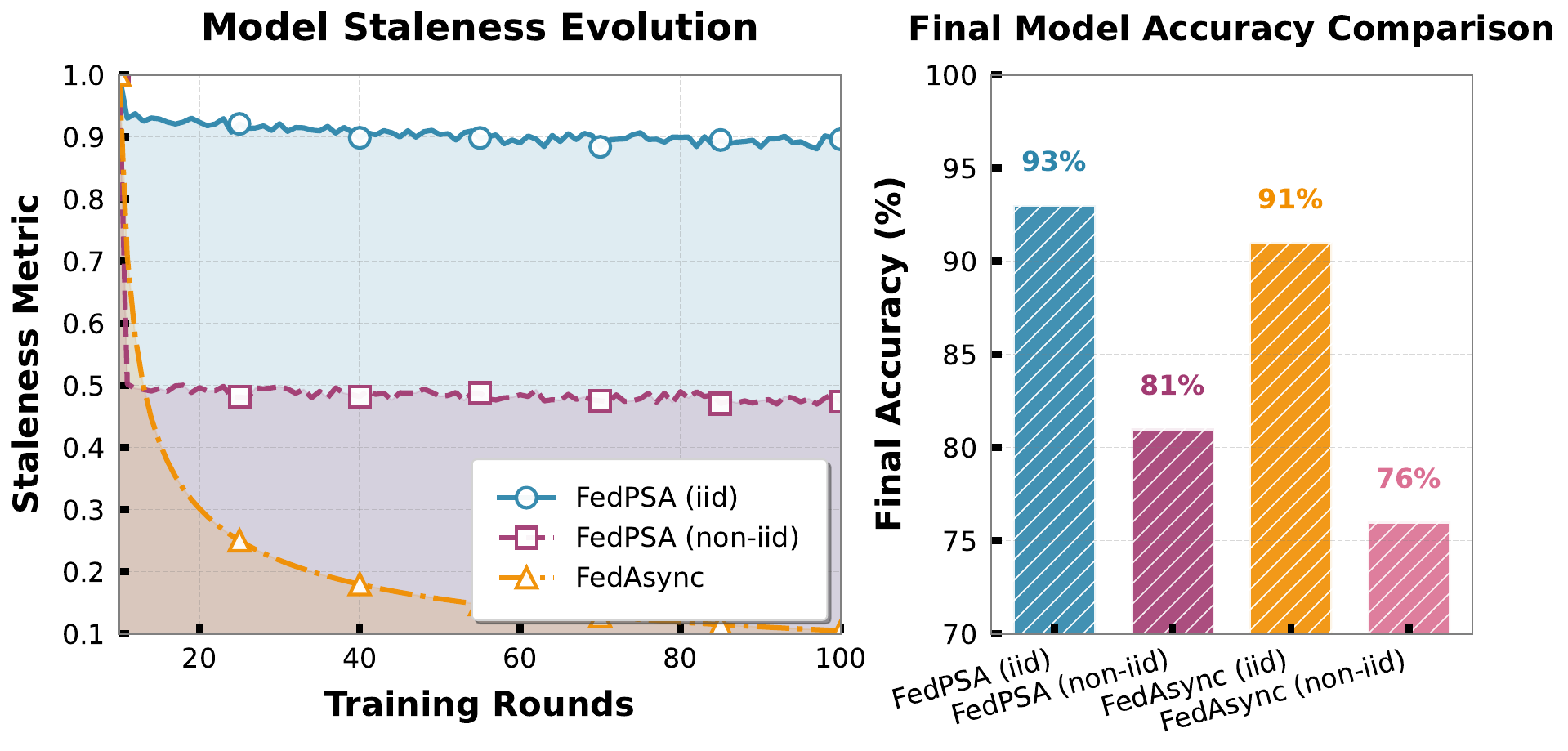} 
    \caption{Comparison of weighting coefficients and final accuracy between FedPSA and FedAsync. Traditional methods overlook the details during aggregation, leading to poor final performance.}  
    \label{fig:figure1}
\end{figure}

\begin{figure}[t]
    \centering
    \includegraphics[width=0.5\textwidth]{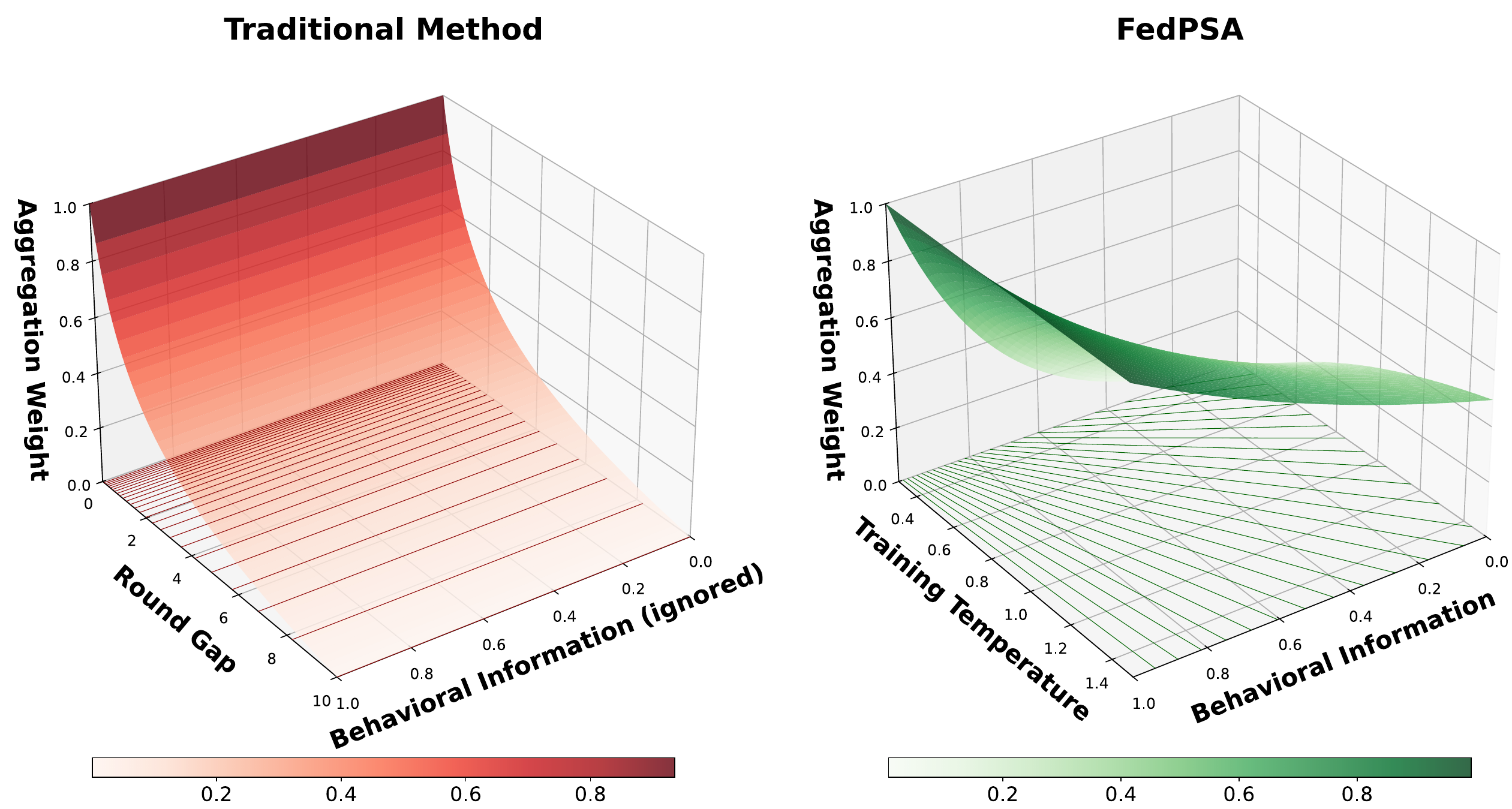} 
    \caption{Weighting schemes in AFL: round gap vs. behavioral information (FedPSA). The traditional method selects $\frac{1}{\sqrt{\tau + 1}}$ as the weighting scheme.}  
    \label{fig:figure2}
\end{figure}

To validate this idea, we conduct a series of comparative experiments to examine how traditional round-gap-based staleness and sensitivity-based staleness behave under different scenarios. The experimental results are illustrated in Figs. \ref{fig:figure1} and \ref{fig:figure2}. As long as the round difference is fixed, traditional metrics always produce identical staleness values, regardless of training stage or data distribution. In contrast, sensitivity-based staleness exhibits the following desirable properties:
\begin{enumerate}
    \item \textbf{Distribution awareness.} When a client update is generated under a large data distribution shift, its sensitivity pattern differs more from the server’s, leading to a lower aggregation weight.
    \item \textbf{Stage awareness.} For the same round gap, updates in the early phase and late phase receive different weights, reflecting the changing dynamics as the model converge.
    \item \textbf{Saturation effect.} As the round gap grows beyond a certain threshold, the measured staleness gradually saturates, rather than increasing unboundedly, which better matches the empirical behavior of asynchronous training.
\end{enumerate}

These observations suggest that parameter sensitivity offers a more reasonable and informative proxy for model staleness, forming the foundation of our proposed aggregation scheme.

\section{Methodology}
\label{sec:format}
In this section, we describe the workflow of FedPSA, an AFL algorithm designed to assist in aggregation by leveraging parameter sensitivity. By down-weighting conflicting updates during aggregation, it can effectively improve the final model performance. Additionally, we provide the convergence analysis of FedPSA in Appendix.
\begin{algorithm}[t]
\caption{FedPSA}
\label{alg:fedpsa}
\resizebox{0.75\linewidth}{!}{%
\begin{minipage}{\linewidth}
\begin{algorithmic}[1]
\Require initial server model $w_g^0$, all clients $c_n$, thermometer queue $Q$, queue length $\mathcal{L}_q$, buffer $S$, buffer size $\mathcal{L}_s$, local epochs $E$, random projection matrix $R$
\State The server broadcast the public batch data $D_b$ and projection matrix $R$ to all clients $c_n$.
\For{each round $t = 1,2,\ldots,T$}
    \State sample available clients $c_{avail}$ from $c_n$
    \State \textbf{Clients Execute:}
    \For{each client $c_i$ in $c_{avail}$ \textbf{parallel} do}
        \State initialize $w_i^0 \leftarrow w_g^{t-1}$
        \State $w_i^{t} \leftarrow \text{LocalUpdate}(w_i^0, D_i)$
        \State $\Delta w_i = w_i^{t} - w_i^0$
        \State $\mathbf{s}_i \leftarrow \text{ComputeSensitivity}(w_i^{t}, D_b)$
        \State $\tilde{\mathbf{s}}_i \leftarrow R\, \mathbf{s}_i$
        \State send $(\Delta w_i, \tilde{\mathbf{s}}_i)$ to Server
    \EndFor
    \State \textbf{Server Executes:}
    \If{receive clients' package}
        \State Receive $(\Delta w_i, \tilde{\mathbf{s}}_i)$ from client $i$

        \State Push tuple($\Delta w_i$,$\kappa_i \leftarrow \text{CosineSimilarity}(\tilde{\mathbf{s}}_i, \tilde{\mathbf{s}}_g)$) into buffer $S$, $m_i \leftarrow \sum_{j=1}^d (\Delta w_i^{(j)})^2$ into queue $Q$
        \If{queue $Q$ not full}
            \State Weighting scheme: uniform averaging
        \ElsIf{queue $Q$ is full for the first time}
            \State $M_0 = \text{Average}(Q)$
        \EndIf
        \State $M_{\text{cur}} = \text{Average}(Q)$
        \State $\text{Temp} = \left(\dfrac{M_{\text{cur}}}{M_0}\right)\cdot \gamma + \delta$
        \If{$\text{length}(S) \ge \mathcal{L}_s$}
            \If{Weighting scheme is null}
                \For{each $i$ in buffer $S$}
                    \State $\text{$\mathrm{Weight}$}_i =
                    \dfrac{\exp\!\left(\kappa_i / \text{$Temp$}\right)}
                          {\sum_{j=1}^{\mathcal{L}_s} \exp\!\left(\kappa_j / \text{$Temp$}\right)}$
                \EndFor
            \EndIf
            \State $w_g^{t} =
                w_g^{t-1}
                + \sum_{i=1}^{\mathcal{L}_s} \text{$\mathrm{Weight}$}_i \cdot \Delta w_i$
            \State clear buffer $S$
        \EndIf
    \EndIf
\EndFor
\Ensure the fully trained server model $w_g^T$
\end{algorithmic}
  \end{minipage}%
}
\end{algorithm}

\subsection{FedPSA Overview}
Based on the aforementioned motivations, we propose FedPSA, a simple yet effective method improved upon FedBuff. This method only adjusts the weight calculation method during the server-side aggregation phase. However, by examining the model staleness issue from a novel perspective, FedPSA successfully establishes a new criterion for assessing model staleness in AFL, thereby further unlocking the potential of asynchronous methods.

To facilitate readers' quick understanding of the FedPSA framework, the following section provides an overview of its algorithmic process. The buffer, as a successful design in AFL, can temporarily store model updates while supporting concurrent client training, significantly reducing the frequency of global model updates and thus mitigating the negative impact of model staleness. Therefore, FedPSA also adopts this design. The core innovation of FedPSA lies in its weighting strategy within the buffer: although it also performs a weighted average of the updates in the buffer based on staleness, its criterion for judging staleness is fundamentally different from traditional methods.

FedPSA requires each client to submit its parameter sensitivity matrix along with the model update when uploading. When the buffer is full, the server compares this matrix with the parameter sensitivity matrix of the global model to generate a corresponding matrix similarity score, which is used to evaluate model staleness. However, this score alone does not directly determine the aggregation weights. FedPSA also incorporates the current training phase for a comprehensive judgment. Specifically, FedPSA exhibits varying degrees of tolerance for model staleness at different training stages: it encourages exploration of more solution spaces in the early stages of training, while promoting stable model convergence in the later stages. To identify the current training phase, FedPSA introduces a queue structure called the training thermometer, which records the momentum information of recent client training. By integrating both the training temperature and model staleness, FedPSA performs a weighted aggregation of the gradients in the buffer. The specific process is outlined in Algorithm \ref{alg:fedpsa}.
\subsection{Model Parameter Sensitivity}
\label{subsec:sensitivity}

We follow the standard pruning literature \cite{DBLP:conf/iclr/LeeAT19,inproceedings} and define the sensitivity of
a model parameter as the change of the loss when this parameter is set
to zero. Consider a model with parameter vector
\(\Theta = \{\theta_1, \theta_2, \dots, \theta_d\}\) and loss function
\(\mathcal{F}(\Theta)\). The sensitivity of the \(i\)-th parameter is
\begin{equation}
    s_i 
    = \bigl| \mathcal{F}(\Theta) - \mathcal{F}(\Theta - \theta_i \mathbf{e}_i) \bigr|,
\end{equation}
where \(\mathbf{e}_i\) denotes the \(i\)-th canonical basis vector.
Computing \(s_i\) exactly for all parameters requires one forward pass
per parameter, which is prohibitive for modern deep models. We therefore approximate using a second-order Taylor expansion.

\paragraph{\textbf{Second-order Taylor approximation}}
We expand the loss at \(\Theta\) along the direction
\(-\theta_i \mathbf{e}_i\) up to second order:
\begin{equation}
\begin{aligned}
    \mathcal{F}(\Theta - \theta_i \mathbf{e}_i) 
    &\approx \mathcal{F}(\Theta)
    + \nabla_i \mathcal{F}(\Theta)(-\theta_i)
    + \frac{1}{2}(-\theta_i) H_{ii}(\Theta)(-\theta_i) \\
    &\quad + R_2 \\
    &= \mathcal{F}(\Theta)
    - \nabla_i \mathcal{F}(\Theta)\,\theta_i
    + \frac{1}{2} H_{ii}(\Theta)\,\theta_i^2 + R_2,
\end{aligned}
\end{equation}
where \(\nabla_i \mathcal{F}(\Theta)\) is the partial derivative
w.r.t.~\(\theta_i\), \(H_{ii}(\Theta)\) is the \(i\)-th diagonal element
of the Hessian \(H(\Theta) = \nabla^2 \mathcal{F}(\Theta)\), and \(R_2\)
is the remainder term. Ignoring \(R_2\), we obtain
\begin{equation}
\begin{aligned}
    s_i 
    &= \bigl| \mathcal{F}(\Theta) - \mathcal{F}(\Theta - \theta_i \mathbf{e}_i) \bigr| \\
    &\approx \Bigl| \nabla_i \mathcal{F}(\Theta)\,\theta_i 
          - \frac{1}{2} H_{ii}(\Theta)\,\theta_i^2 \Bigr|.
\end{aligned}
\label{eq:second_order_sensitivity}
\end{equation}

Equation~\eqref{eq:second_order_sensitivity} is the ideal second-order
sensitivity, but still requires access to the diagonal of the Hessian
and the gradient at \(\Theta\).

\paragraph{\textbf{Approximating the Hessian diagonal by the Fisher information}}
Explicitly forming the Hessian \(H(\Theta)\) costs \(O(n^2)\) and is
intractable for high-dimensional models. For losses that are (or are
close to) negative log-likelihoods and for models near a local optimum,
it is standard to approximate the Hessian by the Fisher information
matrix. Concretely, given a dataset or mini-batch
\(\{x_k\}_{k=1}^m\) and the corresponding per-sample (or mini-batch)
losses \(\mathcal{F}_k(\Theta)\), we define the empirical Fisher
diagonal as
\begin{equation}
    F_{ii}(\Theta)
    = \frac{1}{m} \sum_{k=1}^{m}
      \bigl( \nabla_{\theta_i} \mathcal{F}_k(\Theta) \bigr)^2.
\end{equation}

Under the usual regularity assumptions from information geometry,
\(F_{ii}(\Theta)\) provides an approximation of the expected Hessian
diagonal \(\mathbb{E}[H_{ii}(\Theta)]\). We therefore replace
\(H_{ii}(\Theta)\) in Equation~\eqref{eq:second_order_sensitivity} by
\(F_{ii}(\Theta)\), yielding
\begin{equation}
    s_i \approx 
    \Bigl| \nabla_i \mathcal{F}(\Theta)\,\theta_i 
          - \frac{1}{2} F_{ii}(\Theta)\,\theta_i^2 \Bigr|.
\label{eq:fisher_approx}
\end{equation}

\paragraph{\textbf{Practical sensitivity in FedPSA}}
In FedPSA, we evaluate the above approximation at the current
model \(\Theta^t\) at round \(t\). For the \(i\)-th parameter, the
resulting second-order sensitivity is
\begin{equation}
    s_i^t \approx
    \Bigl| \nabla_i \mathcal{F}(\Theta^t)\,\theta_i^t
          - \frac{1}{2} F_{ii}(\Theta^t)\,(\theta_i^t)^2 \Bigr|.
\label{eq:practical_sensitivity}
\end{equation}

Both the gradient \(\nabla \mathcal{F}(\Theta^t)\) and the Fisher
diagonal \(F_{ii}(\Theta^t)\) are computed on a small calibration
mini-batch that is shared across all clients (see below), so that
sensitivities are directly comparable across clients without leaking private data. Note that
Equation~\eqref{eq:practical_sensitivity} contains no additional
task-specific hyperparameters: it follows directly from the
second-order Taylor expansion and the Fisher approximation.

\paragraph{\textbf{Common calibration data for comparable sensitivities}}
To ensure that sensitivity scores from different clients are comparable
and are not dominated by idiosyncrasies of local data, the server
constructs and samples a small shared calibration mini-batch and
broadcasts it to all participating clients prior to training. This
calibration batch does not need to contain any real user data; in
practice, it can be instantiated entirely by synthetic samples
(\emph{e.g.}, i.i.d. Gaussian noise), so that no additional privacy risk is
introduced beyond standard FL. During
training, each client periodically evaluates the gradients
\(\nabla_{\theta_i} \mathcal{F}_k(\Theta^t)\) and the Fisher diagonals
\(F_{ii}(\Theta^t)\) on this shared calibration batch, and computes its
local sensitivity matrix according to
Equation~\eqref{eq:practical_sensitivity}. These sensitivity matrices are
sent to the server together with the model updates and are then used in
FedPSA's aggregation rule.

\subsection{Why Parameter Sensitivity Instead of Other Signals}
A natural question is whether behavioral staleness in AFL can be measured by other signals beyond parameter sensitivity. We considered three alternatives that are commonly used to characterize client updates: \textbf{(a)} gradient-similarity-based weighting, \textbf{(b)} representation/feature-consistency-based alignment, and \textbf{(c)} client-drift or distance-to-global proxies. However, each of them has a notable drawback in the asynchronous setting.
First, gradient similarity is often computed on stale gradients produced by past model versions; due to the version gap, the similarity can quickly become uninformative or even misleading for the current global model. Second, feature-consistency objectives typically require transmitting or reconstructing intermediate activations, additional auxiliary statistics, or extra forward/backward passes, which increases communication and computation overhead beyond what is desirable in lightweight AFL. Third, drift-based proxies are usually coarse distance surrogates that do not distinguish “far but directionally consistent” updates from “near but directionally conflicting” ones, which is exactly the type of ambiguity that behavioral staleness should resolve.
These limitations motivate our choice of parameter sensitivity patterns evaluated on a shared calibration batch, which provide a model-intrinsic, direction-aware, and communication-efficient signal after sketching.

\subsection{Behavioral Staleness via Sensitivity Sketching}
\label{sec:semantic_staleness}
In this paper, we use the term behavioral information to refer to the behavioral response of a model to a shared calibration batch, captured by its parameter-sensitivity pattern. Thus,``behavioral'' here contrasts with purely time-based quantities, and focuses on how the model functionally behaves on data rather than when the update was generated.

In most existing asynchronous FL methods, the aggregation weight of a
client update is modeled as a function of the version gap between
the client and the server. Let $\tau_i^{\text{time}}$ denote the round difference between the local model of client $i$ and the
latest global model on the server. A typical time-based staleness model
takes the form
\begin{equation}
    \alpha_i
    \;=\;
    \alpha\bigl(\tau_i^{\text{time}}\bigr),
    \label{eq:time_staleness}
\end{equation}
where more recent updates (smaller $\tau_i^{\text{time}}$) receive
larger weights. While convenient, this definition treats staleness as a
purely temporal quantity and ignores how the models themselves
behave. As discussed in Section~\ref{sec:motivation}, the same version
gap can correspond to very different update qualities at different
training stages and under different data distributions.

To move beyond this coarse notion, FedPSA interprets staleness through
the lens of behavioral similarity between models. Intuitively,
instead of asking “how many rounds old is this update”, we ask “how
compatible is this update with the current global training dynamics”.
This behavioral view is realized via the parameter-sensitivity vectors
introduced in Section~\ref{subsec:sensitivity}. Let
$\mathbf{s}_i \in \mathbb{R}^d$ denote the flattened
parameter-sensitivity vector of client $i$ (evaluated on the shared
calibration batch), and let $\mathbf{s}_g \in \mathbb{R}^d$ be the
sensitivity vector of the current global model. In the full parameter
space, their behavioral alignment can be characterized by the cosine
similarity
\begin{equation}
    \cos\!\bigl(\mathbf{s}_i,\mathbf{s}_g\bigr)
    \;=\;
    \frac{\bigl\langle \mathbf{s}_i,\mathbf{s}_g \bigr\rangle}
         {\bigl\|\mathbf{s}_i\bigr\|_2 \,
          \bigl\|\mathbf{s}_g\bigr\|_2}
    \;\in [-1,1],
    \label{eq:full_space_cosine}
\end{equation}
where a larger value indicates that the client and server respond to the
calibration data in a more similar way. Updates with high cosine
similarity are thus considered behaviorally “fresh”, whereas updates
with low or even negative similarity are treated as behaviorally “stale”
because they push the model in a direction that conflicts with the
current global behavior.

In principle, one could directly use the full-space cosine
Equation~\eqref{eq:full_space_cosine} as the behavioral staleness signal.
However, transmitting the full $d$-dimensional sensitivity vector
$\mathbf{s}_i$ from each client in every upload can be costly when the
model contains millions of parameters. To make the behavioral staleness
practical in large-scale settings, FedPSA employs a simple
random-projection-based sensitivity sketch, which preserves the
relevant geometry while significantly reducing communication overhead.

At the beginning of training, the server samples a random projection
matrix $R \in \mathbb{R}^{k \times d}$ with $k \ll d$ (\emph{e.g.}, with
i.i.d.\ entries of zero mean and variance $1/k$) and broadcasts $R$ to
all clients. The same matrix $R$ is reused throughout training. Given
its full sensitivity vector $\mathbf{s}_i$, client $i$ forms a
$k$-dimensional sketch by
\begin{equation}
    \tilde{\mathbf{s}}_i
    \;=\;
    R\,\mathbf{s}_i
    \;\in\;
    \mathbb{R}^k,
    \qquad
    \tilde{\mathbf{s}}_g
    \;=\;
    R\,\mathbf{s}_g
    \;\in\;
    \mathbb{R}^k,
    \label{eq:sensitivity_sketch}
\end{equation}
and only needs to transmit $\tilde{\mathbf{s}}_i$ to the server together
with the model update. On the server side, the behavioral similarity used
by FedPSA is then evaluated directly in the sketch space as
\begin{equation}
    \kappa_i
    \;=\;
    \cos\!\bigl(\tilde{\mathbf{s}}_i,\tilde{\mathbf{s}}_g\bigr)
    \;=\;
    \frac{\bigl\langle \tilde{\mathbf{s}}_i,\tilde{\mathbf{s}}_g \bigr\rangle}
         {\bigl\|\tilde{\mathbf{s}}_i\bigr\|_2 \,
          \bigl\|\tilde{\mathbf{s}}_g\bigr\|_2}.
    \label{eq:semantic_similarity_kappa}
\end{equation}

We use $\kappa_i$ in all subsequent formulas as the behavioral similarity score of client $i$. Conceptually, it can be viewed as a low-dimensional approximation of the full-space cosine in Equation~\eqref{eq:full_space_cosine}.

With this sketching mechanism, the communication cost per client upload
for sensitivity information is reduced from $d$ floating-point entries
to $k$ entries. The effective compression ratio is
\begin{equation}
    \text{compression ratio}
    \;=\;
    \frac{k}{d},
\end{equation}
which can be smaller than $10^{-2}$ when $d$ is on the order of
$10^5$--$10^6$ and $k$ is fixed to a few hundred.

Classical Johnson--Lindenstrauss (JL) results provide a theoretical
justification for using such random projections. Let
$\{x_\ell\}_{\ell=1}^N \subset \mathbb{R}^d$ be a collection of vectors
of interest. There exists a random linear map
$P \in \mathbb{R}^{k \times d}$ with
$k = \mathcal{O}(\log N / \varepsilon^2)$ such that, with high
probability, all pairwise squared distances are preserved up to a factor
$(1 \pm \varepsilon)$:
\begin{equation}
\begin{aligned}
    (1 - \varepsilon)\,\|x - y\|_2^2
    \;\le\;
    \|P x - P y\|_2^2
    \;\le\;
    (1 + \varepsilon)\,\|x - y\|_2^2, 
    \\
    \quad\quad \forall\, x,y \in \{x_\ell\}.
\end{aligned}
\label{eq:jl_distance_preservation}
\end{equation}
For $\ell_2$-normalized vectors, squared distance and cosine similarity
are linked via
\begin{equation}
    \|x - y\|_2^2
    \;=\;
    2 - 2\langle x,y\rangle
    \;=\;
    2\bigl(1 - \cos(x,y)\bigr),
    \label{eq:distance_cosine_relation}
\end{equation}
so the preservation of distances in Equation~\eqref{eq:jl_distance_preservation}
directly implies that the cosine similarity computed in the sketch space
is close to that in the original space.

\subsection{Training Thermometer}
As training proceeds, the global model gradually converges and the scale
of parameter updates typically decreases. Intuitively, in the early
stage the model is still far from a good solution and can benefit from
exploring more diverse directions, whereas in the late stage it becomes
important to emphasize only those updates that are highly consistent
with the current optimization trajectory. To let FedPSA adapt its
tolerance to behavioral staleness across different training phases, we
introduce a simple training thermometer.

Recall that FedPSA measures the behavioral compatibility between client
$i$ and the server by the cosine similarity $\kappa_i$ between their
parameter-sensitivity sketches (Section~\ref{sec:semantic_staleness}).
Larger $\kappa_i$ indicates better alignment and thus lower behavioral
staleness. If we directly feed $\kappa_i$ into a weighting
scheme, the resulting weights would implicitly assume a fixed “temperature”:
the same level of misalignment would be treated equally at the
beginning and near convergence. The training thermometer instead makes
this temperature depend on the recent dynamics of the model updates.

\begin{figure*}[t]
    \centering
    \includegraphics[width=1\textwidth]{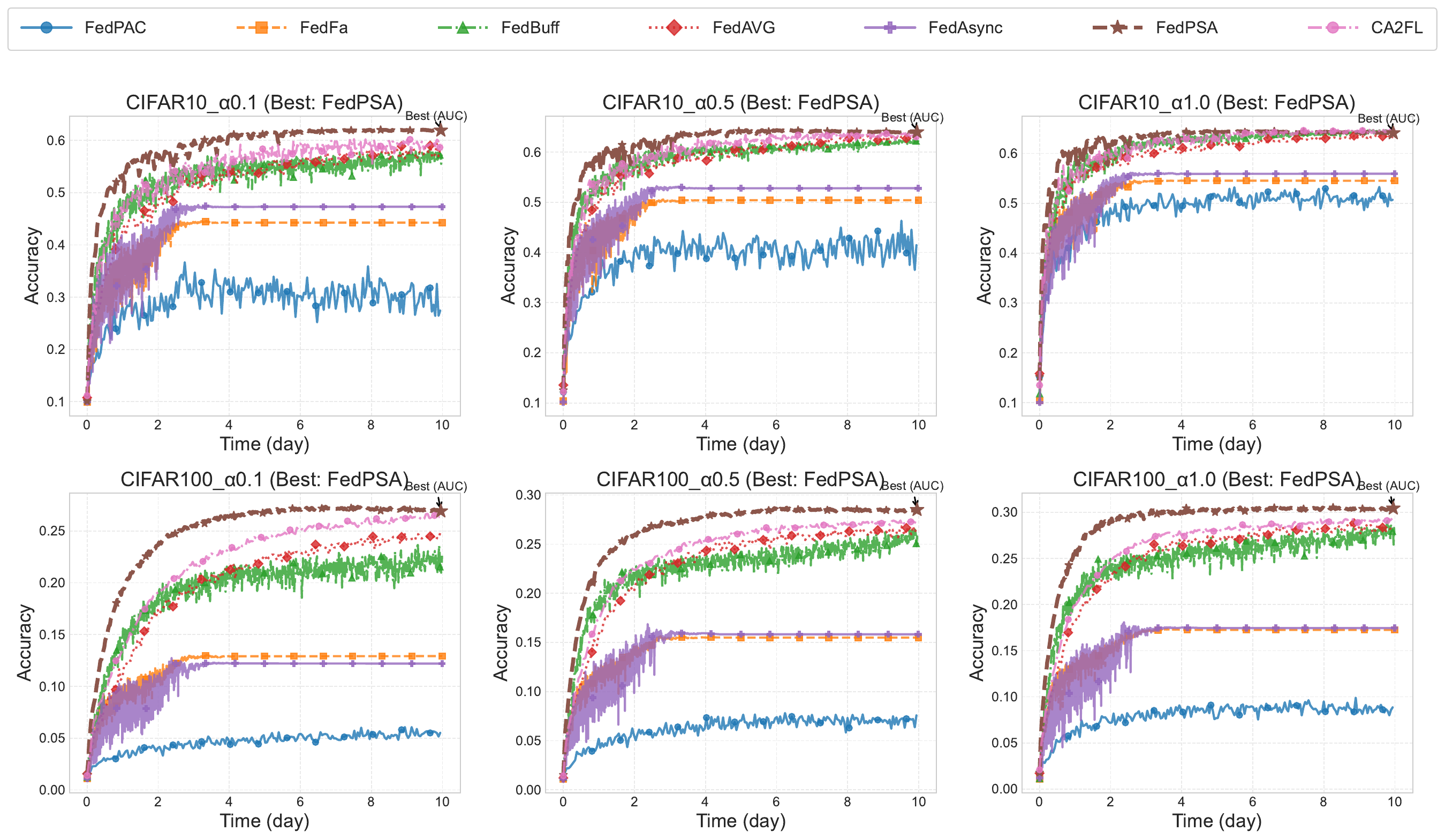} 
    \caption{Convergence curves of different algorithms on the CIFAR dataset.}  
    \label{fig:figure_shoulian}
\end{figure*}

Concretely, the server maintains a fixed-size queue $Q$ to store a
scalar measure of the magnitude of recent client updates. When client
$i$ uploads an update $\Delta w_i$, the server computes its squared
$\ell_2$ norm
\begin{equation}
    m_i
    \;=\;
    \|\Delta w_i\|_2^2
    \;=\;
    \sum_{j=1}^d \bigl(\Delta w_i^{(j)}\bigr)^2,
    \label{eq:update_magnitude}
\end{equation}
and pushes $m_i$ into $Q$. Once the queue is full, the oldest entry is
dropped for each new insertion. Let $M_{\text{cur}}$ denote the average
update magnitude over the current queue,
\begin{equation}
    M_{\text{cur}}
    \;=\;
    \frac{1}{|Q|} \sum_{m \in Q} m,
\end{equation}
and let $M_0$ be the value of this average when the queue is filled for
the first time. We then define the current training temperature as
\begin{equation}
    \mathrm{Temp}
    \;=\;
    \Bigl(\frac{M_{\text{cur}}}{M_0}\Bigr)\,\gamma + \delta,
    \label{eq:temp_def}
\end{equation}
where $\gamma$ and $\delta$ are hyperparameters. In the early stage of
training, the updates are large and $M_{\text{cur}} \approx M_0$, so
$\mathrm{Temp}$ stays relatively high. As the model converges and the
updates become smaller, the ratio $M_{\text{cur}}/M_0$ decreases and
$\mathrm{Temp}$ gradually drops.

Given the behavioral similarity scores $\{\kappa_i\}$ for the updates
stored in the buffer and the current temperature $\mathrm{Temp}$, FedPSA
assigns aggregation weights via a softmax function:
\begin{equation}
    \mathrm{Weight}_i
    \;=\;
    \frac{\exp\bigl(\kappa_i / \mathrm{Temp}\bigr)}
         {\sum_{j=1}^{\mathcal{L}_s} \exp\bigl(\kappa_j / \mathrm{Temp}\bigr)},
    \label{eq:softmax_weights}
\end{equation}
where $\mathcal{L}_s$ is the number of updates in the buffer. A higher
temperature flattens the softmax and thus tolerates a broader range of
behavioral discrepancies, which is desirable in the exploratory early
stage. A lower temperature sharpens the softmax and focuses the
aggregation on a few highly aligned updates, which stabilizes
convergence in the late stage.

Finally, once the buffer is full, the server aggregates the stored
updates according to these weights and applies them to the global model:
\begin{equation}
    w_g^{\text{round}}
    \;=\;
    w_g^{\text{round}-1}
    \;+\;
    \sum_{i=1}^{n_s} \mathrm{Weight}_i \,\Delta w_i.
    \label{eq:fedpsa_update}
\end{equation}

Equations~\eqref{eq:temp_def}–\eqref{eq:fedpsa_update} together realize a temperature-controlled aggregation mechanism: in the early phase, FedPSA allows behaviorally diverse updates to contribute, while in the later phase, it increasingly suppresses behaviorally stale updates and relies mainly on those that are well aligned with the global sensitivity pattern.

\section{Experiment}
\label{sec:experiment}

\subsection{Experimental Setup}
In this section, we use the FLGO \cite{wang2023flgofullycustomizablefederated} framework to compare FedPSA with 6 state-of-the-art baselines on 4 datasets. In the FLGO framework, one virtual day consists of 86,400 atomic time units. All experiments were run on
8 NVIDIA A100-SXM4-80GB GPUs.

\begin{table*}[t]
\centering
\caption{Final accuracy (\%) of MNIST / FMNIST after 10 days of virtual time. Best results are in \textbf{bold}.}
\setlength{\tabcolsep}{5pt}
\renewcommand{\arraystretch}{1.15}
\resizebox{0.5\textwidth}{!}{
\begin{tabular}{l S S S S S S}
\toprule
& \multicolumn{3}{c}{MNIST} & \multicolumn{3}{c}{FMNIST} \\
\cmidrule(lr){2-4}\cmidrule(lr){5-7}
Method
& {$\alpha{=}0.1$} & {$\alpha{=}0.5$} & {$\alpha{=}1.0$}
& {$\alpha{=}0.1$} & {$\alpha{=}0.5$} & {$\alpha{=}1.0$} \\
\midrule
FedBuff (Base) & 98.66 & 98.76 & 98.70 & 84.03 & 83.87 & 84.06 \\
\addlinespace[2pt]
FedAvg   & 98.35 & 98.67 & 98.70 & 80.77 & 78.38 & 82.46 \\
FedAsync & 95.62 & 97.59 & 97.95 & 82.59 & 82.82 & 83.00 \\
CA2FL    & 98.49 & 98.73 & 98.79 & 83.26 & 83.60 & 83.78 \\
FedFa    & 94.72 & 97.12 & 97.75 & 82.19 & 82.73 & 83.25 \\
FedPAC   & 85.75 & 96.46 & 97.23 & \multicolumn{3}{c}{--} \\
\midrule
\rowcolor{gray!10}
FedPSA (Ours) & \textbf{98.53} & \textbf{98.79} & \textbf{98.84}
              & \textbf{83.84} & \textbf{83.93} & \textbf{84.14} \\
\bottomrule
\end{tabular}}
\label{tab:mnist_fmnist} % 可自定义标签，方便引用
\end{table*}

\begin{table*}[t]
\centering
\caption{Final accuracy (\%) of CIFAR-10 / CIFAR-100 after 10 days of virtual time. Best results are in \textbf{bold}.}
\setlength{\tabcolsep}{5pt}
\renewcommand{\arraystretch}{1.15}
\resizebox{0.5\textwidth}{!}{
\begin{tabular}{l S S S S S S}
\toprule
& \multicolumn{3}{c}{CIFAR-10} & \multicolumn{3}{c}{CIFAR-100} \\
\cmidrule(lr){2-4}\cmidrule(lr){5-7}
Method
& {$\alpha{=}0.1$} & {$\alpha{=}0.5$} & {$\alpha{=}1.0$}
& {$\alpha{=}0.1$} & {$\alpha{=}0.5$} & {$\alpha{=}1.0$} \\
\midrule
FedBuff (Base) & 58.09 & 62.60 & 64.50 & 20.33 & 24.19 & 25.97 \\
\addlinespace[2pt]
FedAvg   & 57.09 & 62.86 & 63.22 & 24.08 & 26.07 & 27.98 \\
FedAsync & 47.30 & 52.80 & 55.93 & 12.20 & 15.83 & 17.45 \\
CA2FL    & 59.90 & 63.81 & 63.95 & 26.61 & 27.77 & 30.01 \\
FedFa    & 44.27 & 50.42 & 54.52 & 12.92 & 15.49 & 17.26 \\
FedPAC   & 18.40 & 36.31 & 48.08 &  5.94 &  7.76 &  8.42 \\
\midrule
\rowcolor{gray!10}
FedPSA (Ours) & \textbf{61.83} & \textbf{63.98} & \textbf{64.04}
              & \textbf{26.70} & \textbf{28.37} & \textbf{30.45} \\
\bottomrule
\end{tabular}}
\label{tab:cifar10_cifar100} % 可自定义标签，方便引用
\end{table*}

\begin{table*}[t]
\centering
\caption{Comparison of AULC after 10 days of virtual time across CIFAR datasets. Best results are in \textbf{bold}.}
\label{tab:aulc}
\setlength{\tabcolsep}{5pt}
\renewcommand{\arraystretch}{1.15}
\resizebox{0.75\textwidth}{!}{
\begin{tabular}{lccccccc}
\toprule
\multirow{2}{*}{\diagbox{Method}{Dataset}} & \multicolumn{3}{c}{CIFAR-10} & \multicolumn{3}{c}{CIFAR-100} & \multirow{2}{*}{Avg} \\ 
\cline{2-7}
& $\alpha=0.1$ & $\alpha=0.5$ & $\alpha=1.0$ & $\alpha=0.1$ & $\alpha=0.5$ & $\alpha=1.0$ & \\ 
\hline
FedBuff (Base) & 5.255 & 5.848 & 6.098 & 1.930 & 2.241 & 2.460 & 3.972 \\
FedAvg & 5.149 & 5.764 & 5.934 & 1.996 & 2.276 & 2.490 & 3.934 \\
FedAsync & 4.431 & 5.038 & 5.351 & 1.109 & 1.451 & 1.600 & 3.163 \\
CA2FL & 5.424 & 5.943 & 6.111 & 2.192 & 2.387 & 2.595 & 4.108 \\
FedFa & 4.187 & 4.841 & 5.269 & 1.194 & 1.447 & 1.614 & 3.092 \\
FedPAC & 2.919 & 3.890 & 4.842 & 0.460 & 0.619 & 0.792 & 2.253 \\ 
\hline
\rowcolor{gray!10} FedPSA (Ours) & \textbf{5.859} & \textbf{6.181} & \textbf{6.246} & \textbf{2.472} & \textbf{2.646} & \textbf{2.871} & \textbf{4.379} \\
\bottomrule
\end{tabular}}
\end{table*}

\paragraph{\textbf{Baselines}} To validate the effectiveness of our proposed FedPSA method, we conducted comparative experiments against six state-of-the-art baselines: FedAvg \cite{pmlr-v54-mcmahan17a}, FedAsync \cite{xie2020asynchronousfederatedoptimization}, FedBuff \cite{nguyen2022federatedlearningbufferedasynchronous}, CA2FL \cite{wang2024tackling}, FedFa \cite{10.24963/ijcai.2024/584}, and FedPAC \cite{xu2023personalizedfederatedlearningfeature}. FedAvg and FedAsync represent pioneering works in synchronous and AFL, respectively, and hold irreplaceable significance in the field. FedBuff introduced the concept of a buffer in AFL for the first time, substantially enhancing performance. CA2FL calibrates global updates by caching the latest client updates. FedFa designs the buffer as a queue structure to enable fully asynchronous training. FedPAC utilizes global behavioral knowledge for feature alignment to learn better representations. For any method-specific hyperparameters, we adhered to the values recommended in the original publications.

\paragraph{\textbf{Datasets}} To evaluate the performance of different methods, we conducted experiments on four commonly used benchmark datasets in FL, including MNIST \cite{lecun2010mnist}, FMNIST \cite{Xiao2017FashionMNISTAN}, CIFAR-10 \cite{Krizhevsky09learningmultiple}, and CIFAR-100 \cite{Krizhevsky09learningmultiple}. We divided each dataset into a test set comprising 10\% of the data and a training set comprising the remainder. To simulate data heterogeneity, we adopted the Dirichlet distribution, widely used in FL, with the parameter $\alpha$ set to 0.1, 0.5, and 1.0. Here, a smaller $\alpha$ value corresponds to a higher degree of data heterogeneity.

\paragraph{\textbf{Network Architectures}}To ensure a comprehensive evaluation, we employ distinct network architectures for different datasets, following standard practices in the field.

For the \textbf{MNIST dataset,} we utilize a Convolutional Neural Network (CNN) \cite{oshea2015introductionconvolutionalneuralnetworks}. The model comprises two 5$\times$5 convolutional layers (the first with 32 channels, the second with 64 channels, each followed by a ReLU activation and a 2$\times$2 max-pooling layer), a flatten layer, and a fully-connected layer with 512 units and ReLU activation \cite{agarap2019deeplearningusingrectified}. The final classification is performed by a linear output layer with 10 dimensions.

For the \textbf{FMNIST dataset,} we adopt a simple linear model. This model consists of a single fully-connected layer that maps the 784-dimensional input directly to a 10-dimensional output, with its bias terms initialized to zero.

For the \textbf{CIFAR-10 dataset,} we employ a deeper CNN. This network features two 5$\times$5 convolutional layers (both with 64 channels, each followed by a ReLU activation and a 2$\times$2 max-pooling layer), followed by a flatten layer and two fully-connected layers (the first with 384 output units and the second with 192 output units, both using ReLU activation). The results are obtained via a final linear output layer with 10 dimensions.

For the \textbf{CIFAR-100 dataset,} we use the same CNN backbone architecture as for CIFAR-10. To accommodate the larger number of classes, the final output layer is adjusted to have 100 dimensions.

\paragraph{\textbf{Parameter Settings}} To ensure a fair comparison between methods, the common hyperparameters of all methods remain consistent. The sampling rate for synchronous FL and the concurrency rate for AFL are both set to 20\%. The initial learning rate is 0.01, with a learning rate decay of 0.999, meaning that after each round, the learning rate decreases to 99.9\% of its previous value. The local epochs for clients are set to 5, batch sizes are set to 64, and SGD is used as the optimizer.
The number of clients is set to 50, with each client's response time following a uniform distribution between 10 and 500 virtual time units. $\mathcal{L}_s$ is set to 5, $\mathcal{L}_q$ is set to 50, $\gamma$ is set to 5, $\delta$ is set to 0.5, and the compressed dimension $k$ is set to 16.

\subsection{Performance Analysis}
\paragraph{\textbf{Final Accuracy}}
This section compares the proposed FedPSA method with various baseline methods. Table~\ref{tab:mnist_fmnist} and Table~\ref{tab:cifar10_cifar100} present the final accuracy of all methods across different datasets, since the model used for the FMNIST task contains only a single fully connected layer, the hierarchical model-based method FedPAC cannot be tested on this task. The experimental results demonstrate that FedPSA achieves the best performance under various datasets and different heterogeneity settings of the same dataset, validating the effectiveness of our method. In particular, on the most challenging dataset, CIFAR-100, FedPSA shows a significant improvement in final accuracy compared to the baseline method FedBuff, further highlighting its capability to effectively mitigate model conflicts in high-difficulty tasks.
\paragraph{\textbf{Convergence Efficiency}}
To evaluate the performance of the proposed method, we compared the convergence efficiency of FedPSA with various existing methods. As shown in Fig. \ref{fig:figure_shoulian}, we plot the learning curves of different methods on two more challenging CIFAR datasets. The convergence trajectory of FedPSA consistently remains above those of other methods. To further quantify the convergence performance of each method, we integrate the learning curves in the figure and derive the Area Under the Learning Curve (AULC) metric, with the specific results summarized in Table~\ref{tab:aulc}.

\paragraph{\textbf{Robustness to System Heterogeneity}}
FedPSA shifts the criterion for staleness evaluation from round differences to the behavioral information of the model. Theoretically, under higher system latency, this method should demonstrate superior performance compared to traditional methods. To evaluate the robustness of FedPSA against system latency, we designed six different system heterogeneity settings, where client response times follow the following distributions: a uniform distribution of 10-500, a long-tail distribution of 10-500, a uniform distribution of 20-1000, a long-tail distribution of 20-1000, a uniform distribution of 50-2500, and a long-tail distribution of 50-2500. Experiments were conducted on the CIFAR-100 dataset, and the results are shown in Table~\ref{tab:sys_heterogeneity_accuracy}. Due to the nature of the long-tail distributions, where most response times cluster around 10, the performance differences among all methods under various long-tail settings are not significant. In the uniform distribution scenarios, FedPSA maintains nearly consistent performance even when client response times double. Moreover, when the response time increases to five times the original, the accuracy drops by only 1.19\%, which is considerably less than the decreases observed in FedBuff (2.36\%) and CA2FL (2.37\%). These results fully demonstrate that FedPSA exhibits stronger robustness in handling system heterogeneity.

\subsection{Hyperparameter Analysis}
\begin{figure}[t]
    \centering
    \includegraphics[width=0.75\columnwidth]{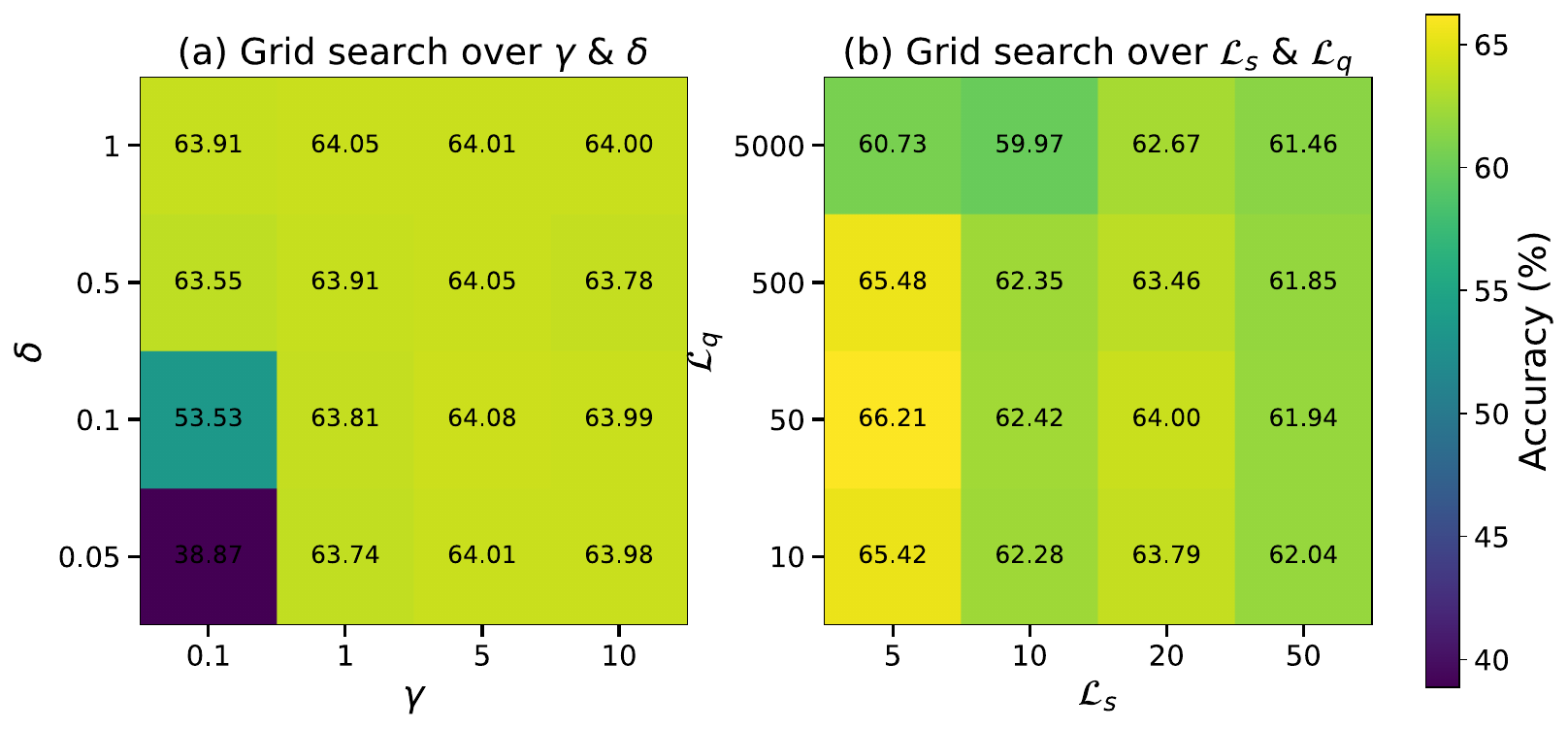} 
    \caption{Performance of FedPSA under different hyperparameters.}  
    \label{fig:figure_hyper}
\end{figure}

To investigate the impact of different hyper-parameters on FedPSA, we conducted a grid-search experiment, the results are shown in Figure \ref{fig:figure_hyper}.
\paragraph{\textbf{Hyperparameter $\gamma$ and $\delta$}}
The parameters $\gamma$ and $\delta$ are used to adjust the coefficients in the temperature calculation. As can be observed from Figure \ref{fig:figure_hyper}, when both hyperparameters are set to relatively small values, the performance of FedPSA drops significantly, whereas in other value ranges, the performance exhibits only minor fluctuations. Therefore, it is not recommended to set both $\gamma$ and $\delta$ to excessively small values simultaneously.
\paragraph{\textbf{Buffer Length $\mathcal{L}_s$ and Queue Length $\mathcal{L}_q$}}
The hyperparameters $\mathcal{L}_s$ and $\mathcal{L}_q$ control the size of the buffer and the length of the stored momentum queue in FedPSA, respectively. When $\mathcal{L}_q$ is set too large, the model struggles to accurately identify the current training phase, leading to performance degradation. Thus, it is advisable to set its value between 10 and 50. If $\mathcal{L}_s$ is assigned a relatively large value, the update frequency of the model decreases, which also adversely affects performance. Therefore, it is recommended to set $\mathcal{L}_s$ within the range of 5 to 20.

\begin{table}[t]
  \centering
  \caption{Final accuracy (\%) of different algorithms under various system heterogeneity settings. \textbf{Bold} indicates the best.}
  \label{tab:sys_heterogeneity_accuracy}
  \setlength{\tabcolsep}{6pt}
  \renewcommand{\arraystretch}{1.15}
  \resizebox{0.5\textwidth}{!}{
  \begin{tabular}{llSS}
    \toprule
    \multirow{2}{*}{Setting} & \multirow{2}{*}{Algorithm} & \multicolumn{2}{c}{Accuracy (\%)} \\
    \cmidrule(lr){3-4}
    & & {Uniform} & {Long-tail} \\
    \midrule

    \multirow{3}{*}{10--500}
      & FedBuff          & 25.75 & 30.23 \\
      & CA2FL            & 28.57 & 30.66 \\
      & \textbf{FedPSA (Ours)} & \textbf{30.23} & \textbf{31.05} \\
    \midrule

    \multirow{3}{*}{20--1000}
      & FedBuff          & 24.92 & 29.79 \\
      & CA2FL            & 27.59 & 30.94 \\
      & \textbf{FedPSA (Ours)} & \textbf{30.21} & \textbf{30.98} \\
    \midrule

    \multirow{3}{*}{50--2500}
      & FedBuff          & 23.39 & 29.89 \\
      & CA2FL            & 26.20 & 29.23 \\
      & \textbf{FedPSA (Ours)} & \textbf{29.04} & \textbf{30.70} \\
    \bottomrule
  \end{tabular}}
\end{table}

\subsection{Ablation Study}
\paragraph{\textbf{Component}}
To evaluate the effectiveness of the FedPSA framework, we conduct three ablation studies:
(1) \textbf{w/o T} (without temperature control): the thermometer mechanism is removed, so gradient weights in the buffer no longer adapt to the current training stage.  
(2) \textbf{w/o S} (without parameter-sensitivity matrix): raw model parameters are used instead of the parameter-sensitivity matrix to measure behavioral similarity.  
(3) \textbf{w/o T\&S} (without both): both temperature control and the parameter-sensitivity matrix are eliminated.

As shown in Table~\ref{tab:abla}, under the Non-IID scenario, removing any component of FedPSA leads to a significant drop in accuracy. This indicates that FedPSA exhibits strong robustness in Non-IID settings due to its increased attention to model-specific details. Moreover, replacing parameter sensitivity with the model itself results in performance degradation, which suggests that parameter sensitivity better captures the behavioral information of the model during training.
\paragraph{\textbf{Calibration Batch Choice}}
A key design choice in FedPSA is the construction of the shared calibration batch $D_b$, which is used solely to compute parameter sensitivity and the corresponding behavioral similarity scores, rather than to train the model itself. To assess how sensitive FedPSA is to the specific choice of $D_b$, we conduct an ablation study comparing two configurations: (i) a calibration batch sampled from the original training data distribution, and (ii) a purely synthetic calibration batch whose inputs are drawn from an i.i.d.\ Gaussian distribution with zero mean and unit variance. In both cases, the size of $D_b$ and the overall training protocol are kept identical.

The results in Table~\ref{tab:calibration_ablation} show that replacing real data with Gaussian noise in $D_b$ leads to negligible differences in final test accuracy and convergence speed across all datasets and asynchronous settings. This observation is consistent with the role of $D_b$ in FedPSA: the algorithm only relies on the relative geometry between the sensitivity vectors of client and server models, which is subsequently smoothed by mini-batch averaging and random projection into low-dimensional sketches. As long as all models are evaluated on the same shared batch, even a simple noise-based $D_b$ provides a sufficiently informative probe to distinguish behaviorally aligned and misaligned updates. Practically, this robustness implies that FedPSA does not require access to additional public or de-identified task data for calibration; instead, one can construct $D_b$ from synthetic noise without sacrificing performance, which further reduces the risk of privacy leakage associated with sharing calibration data.

\begin{figure*}[t]
    \centering
    \includegraphics[width=1\textwidth]{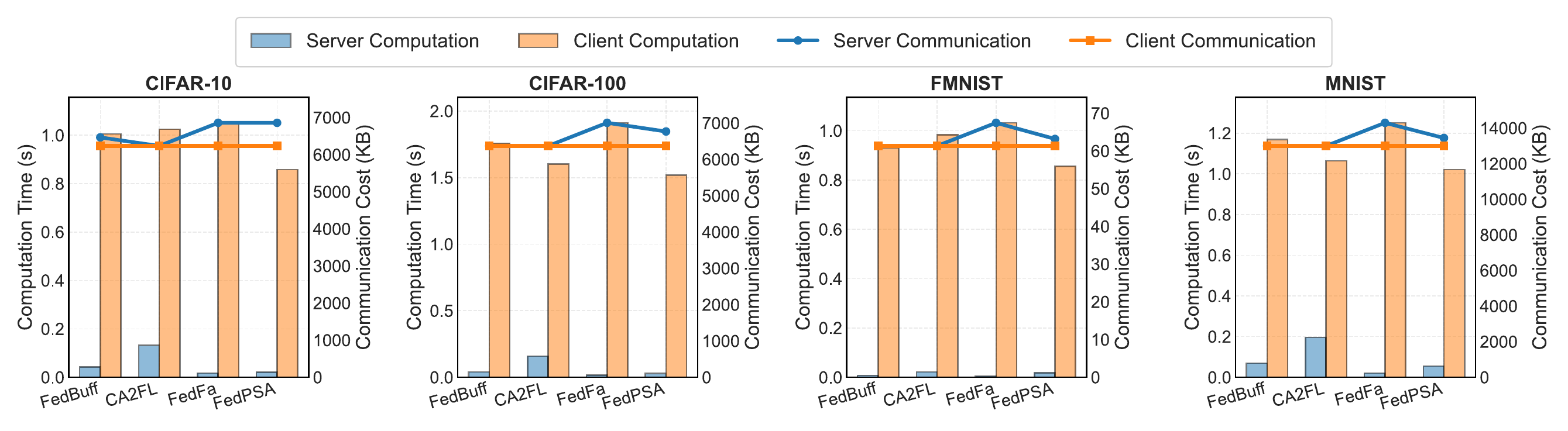} 
    \caption{Computational and communication overheads of different methods on diverse datasets.}  
    \label{fig:overhead}
\end{figure*}

\begin{table}[t]
    \centering
    \caption{Ablation on the calibration batch $D_b$ on CIFAR-10 with different Dirichlet partition parameters. 
    We report final test accuracy (\%) for real-data and Gaussian calibration batches under various batch sizes.}
    \label{tab:calibration_ablation}
    \resizebox{0.65\linewidth}{!}{
    \begin{tabular}{c c ccc}
        \toprule
        Dataset & Batch size 
        & Real-data $D_b$ & Gaussian $D_b$ & Abs.\ $\Delta$ \\
        \midrule
        \multirow{4}{*}{CIFAR-10 ($\alpha=0.1$)}
            & 16   & 63.30 & 63.68 & -0.38 \\
            & 32   & 63.91 & 63.67 & +0.24 \\
            & 128  & 63.84 & 63.24 & +0.60 \\
            & 512  & 63.76 & 63.61 & +0.15 \\
        \midrule
        \multirow{4}{*}{CIFAR-10 ($\alpha=1.0$)}
            & 16   & 66.15 & 66.11 & +0.04 \\
            & 32   & 66.16 & 66.35 & -0.19 \\
            & 128  & 66.42 & 66.42 & +0.00 \\
            & 512  & 66.13 & 66.65 & -0.52 \\
        \bottomrule
    \end{tabular}
    }
\end{table}

\begin{table}[t]
    \centering
    \caption{Ablation Study of FedPSA under Different Settings. \textbf{Bold text} represents the highest value.}
    \resizebox{0.5\textwidth}{!}{
    \begin{tabular}{c c c c c}
        \toprule
        \multirow{2}{*}{\makecell{Dirichlet\\Setting}} & \multirow{2}{*}{Method} & \multicolumn{3}{c}{Concurrency Rate $p$} \\
        \cmidrule(lr){3-5}
        & & $p=0.1$ & $p=0.2$ & $p=0.3$ \\
        \midrule
        \multirow{4}{*}{IID ($\alpha=1$)}
        & w/o T & 62.8 & 63.54 & 64.98 \\ 
        & w/o S & 61.43 & 61.77 & 63.97 \\ 
        & w/o T\&S & 61.51 & 61.68 & 62.74 \\ 
        & \textbf{Full} & \textbf{63.22} & \textbf{64.30} & \textbf{65.76} \\ 
        \midrule
        \multirow{4}{*}{NIID ($\alpha=0.1$)}
        & w/o T & 53.30 & 56.90 & 59.15 \\
        & w/o S & 54.07 & 56.71 & 58.56 \\ 
        & w/o T\&S & 44.08 & 51.09 & 55.16 \\ 
        & \textbf{Full} & \textbf{59.78} & \textbf{61.22} & \textbf{62.39} \\ 
        \bottomrule
    \end{tabular}}
    \label{tab:abla}
\end{table}

\subsection{Communication and Computation Overhead}

Although FedPSA introduces additional parameter sensitivity into the aggregation process, its communication and computation costs remain comparable to existing asynchronous baselines. To verify this, we profile both server-side and client-side computation, as well as the total communication volume, on all four datasets. The aggregated results are summarized in Fig.~\ref{fig:overhead}, where the bars denote computation cost and the line denotes communication cost.

From Fig.~\ref{fig:overhead} we observe that the costs of FedPSA and baselines  are on the same order of magnitude across all datasets. In particular, the client-side communication of FedPSA is almost identical to that of FedBuff, because all methods transmit the same model parameters during each upload. The additional behavioral information accounts for only a very small fraction of the total communication volume. On the computation side, the client cost of FedPSA is also very close to, and in some cases slightly lower than, that of the baselines, indicating that the additional sensitivity computation does not introduce a noticeable burden in practice.

This efficiency mainly comes from two design choices in FedPSA. First, the parameter sensitivity is always evaluated on a fixed public mini-batch rather than on the full local dataset. The extra cost is negligible compared to the normal local training over multiple epochs. Second, instead of transmitting the full high-dimensional sensitivity vector, FedPSA applies a random projection to obtain a low-dimensional sensitivity sketch. Since the sketch dimension is much smaller than the number of model parameters, the communication overhead of sending sensitivity information is reduced by orders of magnitude.

These results demonstrate that FedPSA achieves superior accuracy and robustness while incurring only marginal additional overhead compared with existing asynchronous FL methods. In other words, practitioners do not need to worry about extra communication or computation cost when deploying FedPSA in real-world systems.

\begin{figure}[t]
    \centering
    \includegraphics[width=0.5\columnwidth]{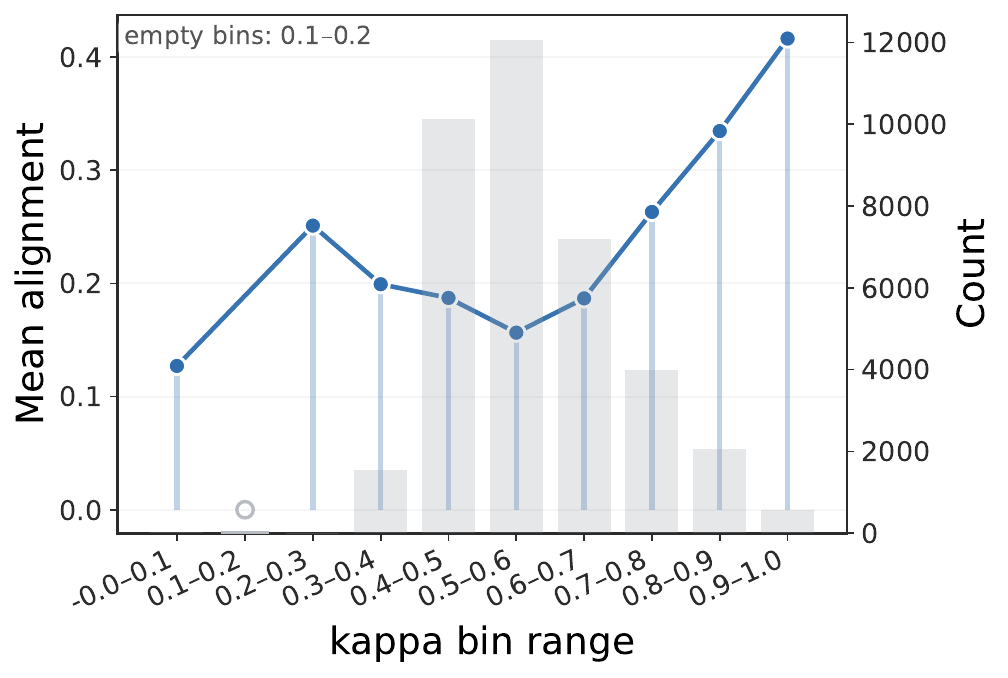} 
    \caption{Relationship between behavioral similarity $\kappa$ and gradient alignment. ``Count'' denotes the number of samples in each bin.}  
    \label{fig:kappa}
\end{figure}

\subsection{Validation of $\kappa$ as a Behavioral Staleness Indicator}
\label{sec:kappa_validation}

In this subsection, we empirically examine whether the proposed
behavioral similarity score $\kappa$ can effectively reflect
\emph{behavioral staleness} in asynchronous federated learning.
Intuitively, if a client update is behaviorally stale with respect to
the current global model, its update direction is more likely to
conflict with the server-side optimization direction. We therefore
study the relationship between $\kappa$ and the alignment between client
updates and the server gradient.

\paragraph{Experimental protocol}
To characterize the behavioral consistency between the client update and the server, we compute the gradients of the global objective on a held-out test batch for both the client and the server:

\begin{equation}
g_{\text{client}} = \nabla_w \mathcal{L}(w_{\text{client}}; D_{\text{test}}), \quad g_{\text{server}} = \nabla_w \mathcal{L}(w_{\text{server}}; D_{\text{test}}).
\end{equation}

We then measure the directional alignment between the client and server gradients as the cosine similarity:

\begin{equation}
\text{align}_i = \cos(g_{\text{client}}, g_{\text{server}}).
\end{equation}

A higher alignment indicates that the client update direction is more consistent with the current server-side optimization trajectory, while a lower alignment suggests stronger behavioral conflict.

We record pairs $(\kappa_i, \text{align}_i)$ for all received updates
throughout training. To reduce the influence of high variance induced by
client heterogeneity and asynchronous delays, we further group updates
into bins according to $\kappa$ (with bin width $0.1$) and compute the
average alignment within each bin.

\paragraph{Correlation analysis}
We first report the correlation between $\kappa$ and alignment at the
sample level. The Pearson and Spearman correlation coefficients are
\[
\text{Pearson } r = 0.1908, \quad
\text{Spearman } \rho = 0.1338.
\]
The relatively weak sample-wise correlation is expected in asynchronous
federated learning, where individual client updates exhibit substantial
variance due to heterogeneous data distributions, different local
training stages, and varying staleness levels.

More importantly, behavioral staleness is inherently a population-level
property that affects aggregation rather than individual updates. We
therefore examine the relationship between $\kappa$ and alignment in
expectation by analyzing binned averages. After binning updates by
$\kappa$, a strong positive and monotonic relationship emerges:
\[
\text{Pearson } r = 0.7620, \quad
\text{Spearman } \rho = 0.6833.
\]
This indicates that updates with larger $\kappa$ values are, on average,
much more aligned with the server gradient, whereas updates with smaller
$\kappa$ tend to exhibit stronger directional inconsistency.
\paragraph{Results and implications}
Figure~\ref{fig:kappa} visualizes the mean alignment as a
function of $\kappa$ bins, together with the number of samples in each
bin. Overall, the mean alignment increases with $\kappa$, indicating
that updates with higher behavioral similarity tend to be more
consistent with the server optimization direction.

We observe that the relationship is not strictly linear, especially in
the low $\kappa$ regime. This phenomenon can be explained by two
factors. First, asynchronous training combined with data heterogeneity
introduces substantial variance, which leads to noisy alignment
measurements at the level of individual updates. Second, several
$\kappa$ bins contain only a limited number of samples, which reduces
the statistical reliability of the corresponding averaged alignment
values. In particular, some low $\kappa$ bins are sparsely populated or
contain no samples at all, and the observed mean alignment in these bins
should therefore be interpreted with caution.

Despite these local fluctuations, the overall monotonic trend remains
clear. Updates with larger $\kappa$ values consistently exhibit higher
average alignment with the server gradient. This result confirms that
$\kappa$ captures meaningful behavioral information related to client
staleness in expectation. Consequently, using $\kappa$ as the core
signal in the aggregation rule of FedPSA is well motivated, as it
enables the server to emphasize behaviorally consistent updates while
suppressing updates that are behaviorally stale.

\section{Conclusion}
\label{sec:conclusion}
In this paper, we propose FedPSA, a novel AFL framework that fundamentally addresses a key limitation of existing asynchronous methods: their reliance on coarse-grained staleness metrics based solely on round or version differences. By introducing parameter sensitivity as a fine-grained, model-intrinsic measure of update staleness, FedPSA directly quantifies the behavioral conflict between client updates and the current global model, thereby enabling more accurate assessment of the actual staleness degree of contributions. Furthermore, we propose a training temperature mechanism implemented through dynamic momentum queues, allowing the framework to adaptively adjust its tolerance for staleness according to the training phase. This mechanism permits broader exploration during early stages when gradient momentum is high, while enforcing stricter convergence behavior in later phases. These two innovations synergize within a buffer-based asynchronous aggregation workflow, significantly mitigating the adverse effects of stale gradients while preserving the efficiency advantages of asynchrony. Extensive experimental results on multiple benchmark datasets and under varying degrees of data heterogeneity demonstrate that FedPSA consistently outperforms state-of-the-art baselines, achieving higher final accuracy and faster effective convergence in the presence of straggler issues.

\section*{ACKNOWLEDGMENTS}
This work was supported in part by the Open Project Program of State Key Laboratory of Integrated Services Networks of Xidian University under Grant. ISN25-25, the Key Research and Development Program of Shaanxi under Grant No. 2024GX-YBXM-556 and the Shenzhen Science and Technology Program under Grant KJZD20240903104400001.

\bibliographystyle{IEEEtran} 
\bibliography{citation} 

\end{document}